\documentclass[letterpaper]{article} % DO NOT CHANGE THIS
\usepackage{aaai2027}  % DO NOT CHANGE THIS
\usepackage[hyphens]{url}  % DO NOT CHANGE THIS
\usepackage{graphicx} % DO NOT CHANGE THIS
\usepackage{natbib}  % DO NOT CHANGE THIS AND DO NOT ADD ANY OPTIONS TO IT
\usepackage{caption} % DO NOT CHANGE THIS AND DO NOT ADD ANY OPTIONS TO IT
\usepackage{algorithm}
\usepackage{algorithmic}

\usepackage{newfloat}
\usepackage{listings}
\DeclareCaptionStyle{ruled}{labelfont=normalfont,labelsep=colon,strut=off} % DO NOT CHANGE THIS
\floatstyle{ruled}
\newfloat{listing}{tb}{lst}{}
\floatname{listing}{Listing}

\usepackage{booktabs}
\usepackage{amssymb} 
\usepackage{amsmath}

\nocopyright

\begin{document}

\title{Standalone DINOv3 for Remote Sensing Training-Free Open-Vocabulary Semantic Segmentation}
\author{
Changhao Zhao$^1$,
Haoxiang Li$^1$,
Yuke Li$^2$,
Hai Liu$^1$,
LingLin Zeng$^{1,\dagger}$
}

% 机构信息
\affiliations{
$^1$College of Resources and Environment, Huazhong Agricultural University, Wuhan 430070, China \\
\quad\{132, hailiu, l\_h\_x\}@webmail.hzau.edu.cn, zenglinglin@mail.hzau.edu.cn\\
}
% \fi
\maketitle
\begin{abstract}
Remote sensing semantic segmentation is hindered by costly pixel-level annotations, motivating training-free open-vocabulary methods. Recently, The recent release of DINOv3 brings DINO.txt, which equips the standalone DINO backbone with image‑text contrastive learning and thus opens up the possibility of open‑vocabulary segmentation.We propose DinoSplat-OV, a training-free framework that adapts DINOv3 to remote sensing without fine-tuning or additional pretraining. Targeting the dense distribution, multi-scale nature, and large size of remote sensing imagery, we design two core modules. Its Text-aware Laplacian Propagation module(TLP) de-noises patch-level predictions by combining textual semantic affinities with local visual similarity, improving regional consistency while preserving boundaries. Its Gaussian Splatting Upsampling module(GSUP) reconstructs pixel-level features through RGB-guided anisotropic aggregation and test-time optimization. A global-anchor sliding-window strategy further supports large-scale imagery. Experiments on UDD5, DOTA, LoveDA and Vaihingen demonstrate competitive or superior performance over existing training-free methods, effectively filling the gap of DINO-series models in training-free open-vocabulary segmentation and providing a viable new path for further advances in this direction.

\end{abstract}

% AnonymousSubmission2027.tex
\section{Introduction}
Remote sensing imagery plays a critical role in precision agriculture, disaster response, and environmental monitoring. However, the acquisition of remote sensing data and pixel‑level annotation are extremely costly, and the images are characterized by dense objects, varying scales, and huge sizes, making vision foundation models (VFMs) pre‑trained on natural images difficult to apply directly. Extensive prior work has attempted fine‑tuning‑based adaptation, yet still faces generalization bottlenecks.

Open‑vocabulary semantic segmentation, which can recognize categories defined by arbitrary text, has emerged as the most promising direction for remote sensing interpretation. Existing methods mostly adopt CLIP as the backbone and post‑process its logits to eliminate global biases, among which training‑free schemes have achieved notable progress. Nevertheless, such optimization heavily relies on CLIP’s dual‑encoder architecture and modifiable internal modules, and cannot be directly transferred to other VFMs.

The release of DINOv3 and its accompanying text encoder (DINO.txt) opens up the possibility of using DINO alone for open‑vocabulary segmentation. Although DINOv3’s visual features are superior to those of CLIP, its visual backbone is completely frozen under the LiT training paradigm, and we are unable to modify its internal attention or feed‑forward layers as we do with CLIP. This leads to severe noise and blurred boundaries when directly applied to remote sensing imagery.

To address this, we systematically transfer the training free experience from CLIP to DINOv3 and propose \textbf{DinoSplat‑OV}. This model is designed for the large‑scale and high‑density characteristics of remote sensing, with a purely inference‑oriented optimization pipeline: synonym aggregation alleviates text‑image matching fragmentation, Text‑aware Laplacian Propagation (TLP) guides feature alignment, 2D Gaussian Splatting Upsampling (GSUP) reconstructs low‑resolution features to pixel‑level precision, and global‑anchor sliding window supports arbitrary large‑image inference.

The contributions of this paper are threefold:
\begin{itemize}
    \item We systematically analyze the differences between DINO and CLIP, and propose the first training‑free open‑vocabulary segmentation framework for remote sensing based on DINOv3.
    \item We develop two core algorithms tailored to DINO.txt: the Text-aware graph Laplacian Propagation (TLP) for coarse-grained feature optimization, and the 2D Gaussian Splatting-inspired upsampling (GSUP) for accurate pixel-level feature reconstruction.
    \item We validate competitive or even superior performance to the state‑of‑the‑art on UDD5, DOTA, LoveDA, Vaihingen and other datasets, without requiring a pre‑trained upsampler.
\end{itemize}

\section{Related Work}
\subsection {Training-Free Open-Vocabulary Segmentation Paradigms.}
\begin{figure*}[t]
\centering
\includegraphics[width=1\textwidth]{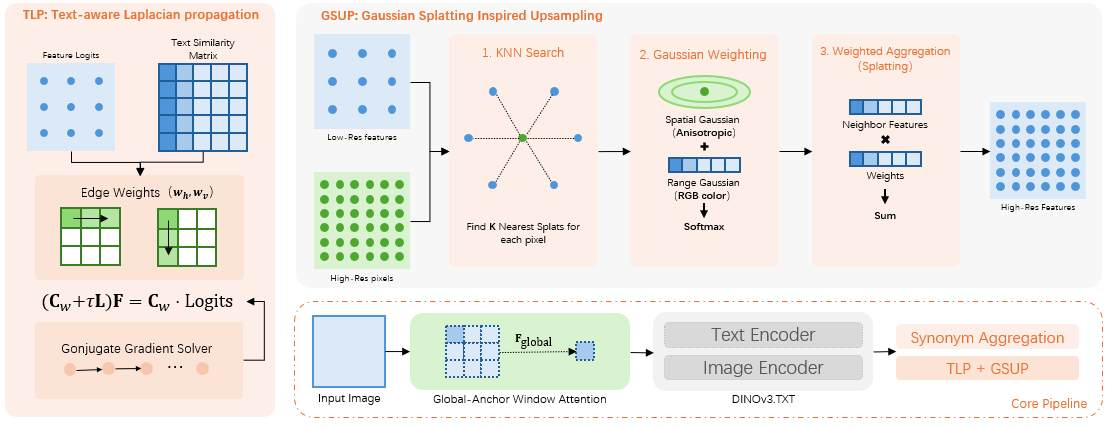} % Reduce the figure size so that it is slightly narrower than the column.
\caption{DinoSplat‑OV employs a sliding‑window strategy to process large remote sensing images; after feature extraction by DINOv3, the features are sequentially refined via Text‑aware Laplacian Propagation (TLP), upsampled through 2D Gaussian Splatting Upsampling (GSUP), and fused with global‑anchor window attention, ultimately producing pixel‑level segmentation predictions.}
\label{fig2}
\end{figure*}
Open-vocabulary segmentation has progressed with vision foundation models (VFMs). CLIP, with its dual-encoder joint training, has become the dominant backbone. Recent training-free efforts, such as SCLIP \citep{wang2023sclip} (modifying self-attention to Query-Query) and ClearCLIP \citep{lan2024clearclip} (removing FFN and residuals), focus on mitigating CLIP's inherent global bias for dense prediction. However, these architectural modifications are intrinsically tied to CLIP's modifiable internal modules.

In parallel, while DINO offers superior visual features, existing works (e.g., clip-dinosier \citep{wysoczanska2023clipdino}, proxyCLIP \citep{lan2024proxyclip}, LPOSS \citep{stojnic2025_lposs}) merely treat it as an auxiliary denoising tool for CLIP, rather than an independent segmenter. The recent release of DINOv3 with its text encoder (DINO.txt) enables DINO-only segmentation for the first time. Critically, unlike CLIP, DINOv3 employs a Locked-image Tuning (LiT) paradigm that fully freezes the visual backbone, making the successful CLIP-style internal modifications inapplicable. This frozen constraint constitutes the core challenge our inference-time optimizations must address.

\subsection {Remote Sensing Adaptations and Domain Gaps.}
Remote sensing imagery presents unique hurdles: dense object distribution, multi-scale targets, and gigapixel resolutions. While OVRS\citep{cao2025open} employs a cost-aggregation training scheme adapted from Cat-Seg\citep{cho2024catseg} to remote sensing data., and SegEarth-OV/SatOV \citep{li2025segearthov} incorporate pre-trained upsamplers (e.g. FeatUp \citep{fu2024featup}) to recover resolution, these methods still rely on dataset-specific pre-training. Consequently, they suffer from degraded generalization when facing unseen geographic regions. Moreover, GLACLIP\citep{lee2026glaclip} introduces a sliding-window strategy tailored for remote sensing, which differs from the standard one. These domain-specific bottlenecks—high-resolution recovery and seamless large-image inference—are exacerbated when using a frozen DINO backbone, as we cannot fine-tune the features to adapt to RS distributions.

\section{Our Method}
\subsection{Preliminaries}

\subsubsection{DINOv3 Text Encoder (DINO.txt)}

\noindent DINO.txt\citep{jose2024dinov2} is the text encoder paired with DINOv3, enabling open‑vocabulary segmentation using DINO alone. Its training paradigm differs fundamentally from CLIP's joint training: DINO.txt adopts a Locked‑image Tuning (LiT) strategy, where the visual backbone is fully frozen and only the text encoder is optimized for cross‑modal alignment. This design preserves the strong discriminative features learned during self‑supervised pre‑training, but consequently prohibits modifying internal attention or feed‑forward layers at inference time. In contrast, CLIP's dual‑encoder architecture allows flexible adjustments to its visual branch, highlighting a core architectural distinction between the two models.

\subsubsection{3D Gaussian Splatting}

\noindent3D Gaussian Splatting (3DGS) \citep{kerbl3Dgaussians} represents continuous visual fields using explicit Gaussian primitives, where each primitive is parameterized by position, covariance, opacity, and appearance attributes. Through differentiable projection and alpha blending, 3DGS reconstructs dense signals from sparse primitives without relying on complex neural decoders. Recent work such as Feat2GS \citep{chen2025feat2gs} further demonstrates that visual foundation model features can be effectively represented by Gaussian primitives. Inspired by this idea, we extend Gaussian splatting from visual rendering to semantic feature reconstruction, treating DINO tokens as semantic Gaussian primitives for high-resolution feature upsampling through adaptive weighted aggregation.

\subsection{Overall Architecture}

The inference pipeline of DinoSplat-OV (Figure 1) proceeds through four sequential modules. 

We first apply synonym aggregation on the text side, fusing embeddings from multiple synonymous descriptions to compensate for DINO.txt's relatively weak cross-modal alignment.

For the frozen visual features, Text-aware Laplacian Propagation (TLP) performs anisotropic diffusion on the initial logits under the guidance of text priors: it aggressively smooths semantically homogeneous regions to suppress noise and fill holes, while preserving discontinuities at land-cover boundaries. The resulting coarse logits are then fed into Gaussian Splatting Upsampling (GSUP), which treats each low-resolution pixel as a 2D Gaussian primitive and reconstructs pixel-level feature maps via RGB-guided anisotropic weighting—all through test-time optimization without any pre-trained upsampler. 

Finally, to handle gigapixel remote sensing inputs, a global-anchor sliding window strategy externally simulates global attention by using the CLS token as contextual reference across windows, coupled with Hann window weighting to eliminate stitching artifacts. This purely inference-oriented design requires neither fine-tuning nor retraining.
\subsection{Synonym Aggregation for Text Embeddings}

To compensate for the weak cross-modal alignment of DINO.txt, we aggregate multiple synonymous descriptions per category. For each class \(c\), let \(\{\mathbf{t}_{c,i}\}_{i=1}^{N_c}\) be the text embeddings of its \(N_c\) synonymous phrases. We compute a unified class embedding via weighted averaging:
\begin{equation}
\mathbf{T}_c = \frac{\sum_{i=1}^{N_c} w_{c,i} \mathbf{t}_{c,i}}{\sum_{i=1}^{N_c} w_{c,i}},
\label{eq:syn_agg}
\end{equation}
where \(w_{c,i}\) are set uniformly in our main experiments, but can be adjusted based on phrase frequency or importance. This aggregation reduces the variance caused by single-word expression biases, providing more stable text priors for subsequent modules.

\subsection{Text-aware Laplacian Propagation (TLP)}
\begin{figure}[htbp]
    \centering
    \includegraphics[width=1\linewidth]{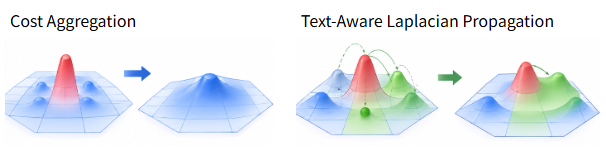}
    \caption{Cost Aggregation vs Laplacian Propagation Diagram}
    \label{fig:placeholder}
\end{figure}
To mitigate noise and cross-window inconsistency in segmentation predictions on low-resolution feature maps, existing post-processing approaches—such as Cost Aggregation~\citep{shou2024cat}—require training on annotated data. To address this limitation, we propose a training-free graph Laplacian propagation algorithm, termed Text-aware Laplacian Propagation (TLP). This method is essentially a variant of label propagation and can be viewed as a closed-form simplification of graph convolutional networks (GCNs) when applied to a single image. Its core idea is to leverage text-embedding priors to construct global semantic relationships among categories and to couple these with local visual features of the image. This coupling guides the classification logits via anisotropic diffusion—smoothing sufficiently within semantically homogeneous regions to eliminate noise, while suppressing cross-region propagation at semantic boundaries to preserve land-cover contours.

Given the aggregated class embeddings \(\mathbf{T} \in \mathbb{R}^{C \times D}\) obtained from Eq.~(1), we first construct a semantic correlation matrix \(\mathbf{S} \in \mathbb{R}^{C \times C}\) via cosine similarity with temperature scaling:
\begin{equation}
S_{ij} = \frac{\exp\left(\mathbf{T}_i^\top \mathbf{T}_j / \tau_S\right)}{\sum_{k=1}^C \exp\left(\mathbf{T}_i^\top \mathbf{T}_k / \tau_S\right)},
\label{eq:semantic_matrix}
\end{equation}
where \(\tau_S\) is a temperature parameter.In practice, we further enhance the diagonal entries and perform row-wise renormalization. In our implementation, we enforce symmetry via \(S \leftarrow (S + S^\top)/2\) to maintain a balanced propagation prior. This matrix serves as a global prior to modulate the subsequent local diffusion intensity.

Let \(P \in \mathbb{R}^{B \times C \times H \times W}\) be the probability map after softmax normalization, and let \(p_c(u)\) denote the probability that position \(u\) belongs to category \(c\). We define the prediction confidence at position \(u\) as \(\gamma(u) = \max_c p_c(u)\), and the semantic self-consistency as
\begin{equation}
\alpha(u) = \sum_{c} p_{c}(u) \sum_{j} S_{cj} p_{j}(u),
\end{equation}
which measures how well the pixel's probability distribution agrees with the text semantic matrix \(\mathbf{S}\)—if the predicted category exhibits consistent responses with semantically similar categories, \(\alpha(u)\) will be high. The diffusion gating coefficient is then defined as:
\begin{equation}
\lambda(u) = \max\{\gamma(u), \gamma_{\min}\}^2 \cdot \bigl(1 + \alpha(u)\bigr),
\label{eq:lambda}
\end{equation}
where \(\gamma_{\min}\) is a small constant (set to \(0.05\) in experiments) to prevent numerical instability. \(\lambda(u)\) controls the diffusion strength: regions with high confidence and strong semantic self-consistency are sufficiently smoothed to eliminate noise, while regions with low confidence or semantic ambiguity suppress diffusion to preserve fine details.

For adjacent pixel positions \(u\) and \(v\), we define the anisotropic edge weight \(\omega_{u,v} \in [0,1]\), which integrates both image- and semantic-guided terms:
\begin{equation}
\omega_{u,v} = \omega_{u,v}^{\mathrm{im}} \cdot \omega_{u,v}^{\mathrm{sem}}, \quad
\hat{\omega}_{u,v} = \frac{\omega_{u,v} + \omega_{v,u}}{2}, 
\label{eq:edge_weight}
\end{equation}
where 
\[
\omega_{u,v}^{\mathrm{img}} = \exp\!\left(-k_I \cdot \frac{|I(u) - I(v)|}{\mu_I}\right),
\]
with \(\mu_I = \mathrm{mean}(|\nabla I|)\) being the mean absolute grayscale gradient over the local map and \(k_I\) a constant (set to \(5.0\) in experiments), and the semantic term is modulated as
\[
\omega_{u,v}^{\mathrm{sem}} = 1 + \sum_{c} p_c(u) \sum_{j} S_{cj} \, p_j(v).
\]
In practice, we apply a lightweight numerical projection 
\(\hat{\omega}_{u,v} \gets \min(1, \hat{\omega}_{u,v})\) 
to strictly confine the symmetric edge weights within \([0,1]\) as a stabilizer.

Let \(X \in \mathbb{R}^{B \times C \times H \times W}\) denote the classification logits to be optimized (initialized as the input logits). Define the graph Laplacian operator \(\mathcal{L}\) acting on a feature map \(Z\) as:
\begin{equation}
\mathcal{L}(Z)(u) = \sum_{v\in \mathcal{N}(u)} \hat{\omega}_{u,v} (Z(u) - Z(v))
\label{eq:laplacian}
\end{equation}
where \(\mathcal{N}(u)\) denotes the four-neighborhood. The final smoothing process is accomplished by solving the following sparse linear system:
\begin{equation}
\lambda(u) X(u) + \tau \cdot \mathcal{L}X(u) = \lambda(u) \cdot X_{\mathrm{in}}(u), \quad \forall u,
\label{eq:linear_system}
\end{equation}
where \(X_{\mathrm{in}}\) is the original input logits and \(\tau > 0\) is a smoothing strength parameter. This formulation is equivalent to performing anisotropic smoothing on low-frequency regions while preserving high-frequency boundaries. The raw edge weight $\omega_{u,v}$ defined in Eq.~(5) is generally asymmetric due to the semantic term. To enable a symmetric graph Laplacian and safely apply the Conjugate Gradient (CG) method, we symmetrize it as $\hat{\omega}_{u,v} = (\omega_{u,v} + \omega_{v,u}) / 2$ (Eq.~(5a)). With this symmetrization, the coefficient matrix in Eq.~(7) becomes symmetric positive definite (since $\lambda(u)>0$ and the graph Laplacian is positive semidefinite), thus guaranteeing the convergence of CG. We therefore employ the CG method for efficient iterative solution, which converges robustly within a fixed number of steps without requiring backpropagation or training.

In practice, to accelerate inference on high-resolution images while maintaining global receptive fields, we downsample the input logits and image to a coarse grid (determined by a target size, e.g., \(72 \times 72\)), solve the linear system in Eq.~\eqref{eq:linear_system} on this downsampled space, and upsample the solution back to the original resolution with GSUP. This downsampling strategy is also consistent with our implementation, where the TLP module operates on the patch tokens' spatial grid before upsampling.

\subsection{GSUP: Gaussian Splatting Inspired Upsampling}
\begin{figure}[htbp]
    \centering
    \includegraphics[width=1\linewidth]{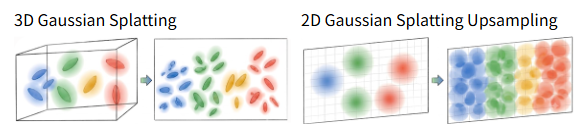}
    \caption{3D Gaussian Splatting vs 2D Gaussian Splatting Upsampling Diagram}
    \label{fig:placeholder}
\end{figure}

For dense segmentation scenarios in remote sensing imagery, recovering low‑resolution features to pixel‑level accuracy is of critical importance. Existing methods such as FeatUp and AnyUp\citep{wimmer2026anyup} rely on pre‑trained upsampling models. However, since their backbones are predominantly trained on natural image datasets like ImageNet, remote sensing data constitutes only a small proportion of their training distribution, leading to notable cross‑domain generalization bottlenecks. Recently, NAF\citep{chambon2025nafzeroshotfeatureupsampling} has improved JBU by replacing its fixed kernel with neighborhood attention, while 3DGS has demonstrated powerful explicit scene representation via anisotropic covariance without pretraining, achieving high‑quality reconstruction through test‑time optimization (TTO). Inspired by both, we propose a feature reconstruction upsampling module based on 2D Gaussian Splatting, termed \textbf{Gaussian Splatting Upsampling (GSUP)}, which similarly replaces JBU's fixed kernel with a Gaussian splatting kernel.
\begin{table*}[t]
  \centering
  \begin{tabular}{l c c c c c c}
    \toprule
    \textbf{Model} & \textbf{Backbone} & \textbf{UDD5} & \textbf{DOTA} & \textbf{LoveDA} & \textbf{Vaihingen} & \textbf{Average} \\
    \midrule
    MaskCLIP \textsubscript{ECCV baseline} & CLIP & 28.1 & 12.4 & 22.6 & 30.0 & 23.3 \\
    DINO.txt \textsubscript{CVPR} & DINOv3 & 32.4 & 17.0 & 27.8 & 35.7 & 28.2 \\
    ClearCLIP \textsubscript{ECCV} & CLIP & 38.2 & 18.5 & 31.6 & 39.4 & 31.9 \\
    LPOSS \textsubscript{CVPR} & CLIP + DINO & 38.8 & 20.2 & 32.4 & 32.0 & 30.8  \\
    SegEarth-OV \textsubscript{CVPR} & ClearCLIP & \textbf{45.3} & 22.3 & \textbf{36.9} & 40.9 & 36.3 \\
    \textbf{DinoSplat-OV\textsubscript{Ours}} & \textbf{DINO.txt} & 42.9 & \textbf{28.6} & 36.3 & \textbf{42.3} & \textbf{37.5} \\
    \bottomrule
  \end{tabular}
  \caption{Comparison of backbones and mIoU for different training free methods.}
  \label{table1}
\end{table*}

\begin{figure*}[t]
  \centering
  \includegraphics[width=1.0\textwidth]{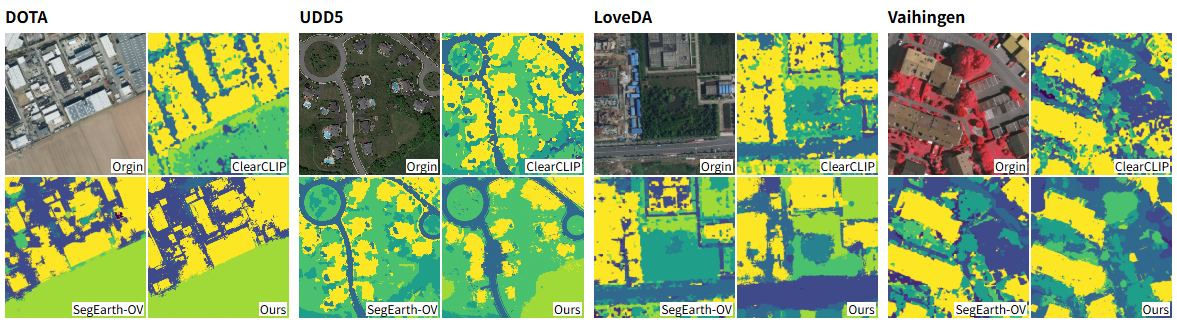} % 宽度相对双栏文本宽度
  \caption{Visualization of segmentation results for different methods on different datasets.}
  \label{fig3}
\end{figure*}
\textbf{Formal Definition.} Let the low-resolution feature map be \(\mathbf{F}_{\mathrm{lr}} \in \mathbb{R}^{B \times C \times H_l \times W_l}\), and the target high-resolution feature map be \(\mathbf{F}_{\mathrm{hr}} \in \mathbb{R}^{B \times C \times H_h \times W_h}\), where \(H_h / H_l = W_h / W_l = r\) is the upsampling factor. We define each pixel \(q \in \mathcal{G}_l\) on the low-resolution feature map (where \(\mathcal{G}_l\) denotes the \(H_l \times W_l\) grid) as a 2D Gaussian splatting primitive, with its coordinate \(\mathbf{x}_q \in \mathbb{R}^2\) serving as the spatial center, the corresponding feature vector \(\mathbf{f}_q \in \mathbb{R}^C\) as the propagable attribute, and the anisotropic covariance matrix \(\boldsymbol{\Sigma}_q \in \mathbb{R}^{2 \times 2}\) characterizing its spatial morphology. The reconstructed feature at a high-resolution target position \(p \in \mathcal{G}_h\) (where \(\mathcal{G}_h\) denotes the \(H_h \times W_h\) grid) is obtained by weighted aggregation of all low-resolution Gaussian primitives:
\begin{equation}
\mathbf{F}_{\mathrm{hr}}(p) = \sum_{q \in \mathcal{N}(p)} w_{p,q} \cdot \mathbf{F}_{\mathrm{lr}}(q),
\end{equation}
where \(\mathcal{N}(p)\) is the set of spatially proximate low-resolution pixels to \(p\) (implemented via \(k\)-nearest neighbor truncation), and \(w_{p,q} \in [0,1]\) are the normalized splatting weights satisfying \(\sum_{q \in \mathcal{N}(p)} w_{p,q} = 1\).

\textbf{Anisotropic Adaptive Weights.} Unlike JBU and similar methods that adopt fixed isotropic kernels, we leverage the current RGB image as guidance to generalize the weights \(w_{p,q}\) to an adaptive anisotropic Gaussian splatting kernel, composed of a spatial term and a color term:
\begin{equation}
w_{p,q} = \operatorname{Softmax}_{q \in \mathcal{N}(p)} \bigl( \phi_s(p,q) + \phi_c(p,q) \bigr),
\end{equation}
where the spatial term employs Mahalanobis distance to measure geometric deformation:
\begin{equation}
\phi_s(p,q) = -\frac{1}{2} \Delta \mathbf{x}_{p,q}^{\top} \boldsymbol{\Sigma}_q^{-1} \Delta \mathbf{x}_{p,q}, \quad \Delta \mathbf{x}_{p,q} = \mathbf{x}_p - \mathbf{x}_q,
\end{equation}
with \(\boldsymbol{\Sigma}_q\) being a learnable Gaussian covariance matrix, parameterized by its scale parameters \(\mathbf{s}_q = (s_x, s_y)\) and rotation angle \(\theta_q\): \(\boldsymbol{\Sigma}_q = \mathbf{R}(\theta_q) \cdot \operatorname{diag}(s_x^2, s_y^2) \cdot \mathbf{R}(\theta_q)^\top\). The color term measures semantic similarity based on RGB pixel differences:
\begin{equation}
\phi_c(p,q) = -\frac{\| \mathbf{I}(p) - \mathbf{I}(q) \|_2^2}{2 \sigma_{r,q}^2},
\end{equation}
where \(\mathbf{I}(p) \in \mathbb{R}^3\) is the RGB color value, and \(\sigma_{r,q}\) is a learnable color bandwidth parameter.

\textbf{Test-Time Optimization (TTO).} In contrast to the feed-forward sampling of traditional JBU, our method treats the Gaussian parameters \(\{\mathbf{s}_q, \theta_q, \sigma_{r,q}\}\) as optimizable variables, takes the low-resolution RGB image \(\mathbf{I}_{\mathrm{lr}}\) as input and the original high-resolution RGB image \(\mathbf{I}_{\mathrm{hr}}\) as supervisory signal, and performs test-time optimization by minimizing the L1 reconstruction loss:
\begin{equation}
\mathcal{L}_{\mathrm{TTO}} = \bigl\| \mathcal{F}_{\mathrm{GSUP}}(\mathbf{I}_{\mathrm{lr}}; \boldsymbol{\Theta}) - \mathbf{I}_{\mathrm{hr}} \bigr\|_1,
\end{equation}
where \(\mathcal{F}_{\mathrm{GSUP}}(\cdot; \boldsymbol{\Theta})\) denotes the forward splatting projection that takes the low-resolution RGB values as the "values", and \(\boldsymbol{\Theta} = \{\mathbf{s}_q, \theta_q, \sigma_{r,q}\}_{q \in \mathcal{G}_l}\). This optimization involves only the Gaussian parameters and does not update any visual backbone or feature encoder. It requires  only a few dozen SGD iterations (e.g., 10 steps) and a sparse neighbor set (K=16), which incurs <3 GB GPU memory and ~1200 ms latency per image, incurring minimal inference overhead. After optimization, with \(\boldsymbol{\Theta}\) fixed, the low-resolution semantic features \(\mathbf{F}_{\mathrm{lr}}\) are fed as the "values" into the same forward pass, yielding the pixel-level high-resolution semantic feature map \(\mathbf{F}_{\mathrm{hr}}\). The entire pipeline requires no pre-training and relies entirely on test-time adaptation, providing a lightweight and generalizable alternative for high-resolution recovery from frozen backbone features.

\subsection{Global-Anchor Window Attention}
Vision foundation models such as CLIP and DINO are trained on images of fixed resolution (e.g., \(224 \times 224\) pixels) and are highly sensitive to input sizes. Remote sensing images, however, typically exceed \(1000 \times 1000\) pixels, making sliding window strategies essential for processing large-scale imagery.

Nevertheless, conventional independent window inference introduces severe stitching artifacts, often resulting in discontinuities at window boundaries and disrupting the spatial consistency of segmentation predictions. To address this, we propose a \textbf{global-anchor guided sliding window attention mechanism} that simulates global self-attention externally, enabling each window to perceive global semantic context during decision-making. Specifically, we adopt the CLS token from ViT as the global anchor \(\mathbf{F}_{\text{global}} \in \mathbb{R}^D\), which aggregates the global representation of the current window during the window encoding process. For the feature \(\mathbf{F}_i \in \mathbb{R}^D\) of the \(i\)-th window, the global-aware weight is computed as:

The global-aware weight for the $i$-th window is computed as:
\[
w_{i} = \frac{\exp\left(\mathrm{sim}(F_i, F_{\mathrm{global}}) / \tau\right)}
        {\sum_{j} \exp\left(\mathrm{sim}(F_j, F_{\mathrm{global}}) / \tau\right)} 
        \cdot G_{\sigma}(i), \tag{13}
\]
where $\mathrm{sim}(\cdot, \cdot)$ denotes cosine similarity, $\tau$ is a temperature coefficient, and $G_{\sigma}(i)$ is a Gaussian window weight defined as $G_{\sigma}(i) = \exp(-d_i^2 / 2\sigma^2)$, with $d_i$ being the normalized distance from the $i$-th window center to the image center. This Gaussian weighting smoothly decays the contribution of windows near the image boundary, effectively suppressing stitching artifacts while preserving the semantic fidelity of central regions. The bandwidth $\sigma$ is set to 0.5 in our experiments to balance boundary smoothness and content preservation.

This mechanism establishes implicit correlations among windows through the global anchor, ensuring that the output of each window is constrained not only by local visual information but also by the global contextual distribution. Meanwhile, the Gaussian weighting \(G_{\sigma}(i)\) effectively eliminates prediction jumps at window seams, enabling seamless end-to-end inference on remote sensing images of arbitrary sizes and generalizing VFMs trained at limited resolutions to large-scale remote sensing scenarios.

\section{Experiments}
As a training‑free framework, DinoSplat‑OV requires no pre‑training or parameter updates and performs end‑to‑end inference directly at test time. Consequently, our experiments focus on performance evaluation and mainly compare against existing training‑free open‑vocabulary segmentation methods.
\subsection{Datasets}
To comprehensively assess the generalisation capability of DinoSplat‑OV across diverse remote sensing scenarios, we select four representative multi‑category segmentation datasets: \textbf{DOTA}\citep{Xia_2018_CVPR} (dense remote sensing objects), \textbf{LoveDA} \citep{4e732ced}(rural‑urban mixed agricultural scenes), \textbf{UDD5}\citep{chen2018large} (drone‑style natural imagery), and \textbf{ISPRS Vaihingen}\citep{rotinwa2012benchmark}(urban scene without blue band). These datasets exhibit significant variations in spatial resolution, object density, and scene complexity, enabling a thorough evaluation of the model's adaptability to various remote sensing image types.
Following SAMRS\citep{SAMRS} , we transform the original DOTA dataset into a semantic segmentation dataset with pixel-level annotations.

\subsection{Comparison with SOTA Methods}
Given that DinoSplat‑OV is designed for training‑free inference, we select the most representative training‑free open‑vocabulary segmentation models as our baselines. To ensure fair comparison, all competing methods are uniformly adapted with a sliding‑window strategy to handle the large input sizes of remote sensing images, with window parameters kept consistent with those of DinoSplat‑OV. Experimental results demonstrate that DinoSplat‑OV achieves overall performance on par with the current state‑of‑the‑art (SOTA), and exhibits particularly pronounced advantages in dense object scenarios (e.g., the DOTA dataset), validating the effectiveness of our approach in remote‑sensing‑specific segmentation tasks.

The experimental results are presented in the figure above. DinoSplat‑OV is built upon DINO.txt, while SegEarth‑OV is based on ClearCLIP. It can be observed that ClearCLIP, by optimizing the CLIP architecture, achieves stronger image‑text matching capability compared to DINO.txt, whereas DINO exhibits superior semantic clustering performance in the visual domain. Building upon this foundation, SegEarth‑OV further introduces a pre‑trained FeatUp upsampling module to attain state‑of‑the‑art performance. In parallel, our approach incorporates Laplacian propagation and a Gaussian‑splatting‑inspired test‑time optimization (TTO) upsampling module. Following a similar optimization trajectory, our method achieves performance on par with the SOTA.

\subsection{GSUP vs Other Feature Reconstruction Models}
To validate the effectiveness of GSUP, we compare it against two representative general‑purpose upsampling models: \textbf{AnyUp}, a generic upsampler that supports DINO features, and \textbf{SatUp}, a dedicated upsampler trained on a remote sensing subset that is also compatible with DINO features. Experimental results demonstrate that \textbf{GSUP}, without any pre‑training, achieves feature reconstruction quality comparable to pre‑trained upsamplers through test‑time optimization (TTO) alone, showing strong competitiveness across multiple remote sensing scenarios.
In our experiments, They reconstruct 72×72 DINOv3 patch features into 224×224 dense feature maps, where each patch token is modeled as an adaptive Gaussian primitive for pixel-level semantic feature recovery.

\begin{table}[htbp]
\centering
\begin{tabular}{l c c c}
  \toprule
  \textbf{Method} & \textbf{Pre Trained} & \textbf{mIoU} & \textbf{Infer Time} \\
  \midrule
  Bilinear\textsubscript{baseline} & No & 35.6 &  0.02s \\
  UPLiFT\textsubscript{CVPR} & $\checkmark$ & 40.6 & 0.1s \\
  AnyUp\textsubscript{ICLR} & $\checkmark$ & 43.8 & 0.3s\\
  SatUp & $\checkmark$(RS) & 43.2 & 0.1s\\
  GSUP\textsubscript{Ours} & TTO & 42.9 & 1.2s\\
  \bottomrule
\end{tabular}
\caption{Performance of different feature upsampling methods on the ImageNet dataset subset.}
\end{table}
\begin{figure}[htbp]
    \centering
    \includegraphics[width=1\linewidth]{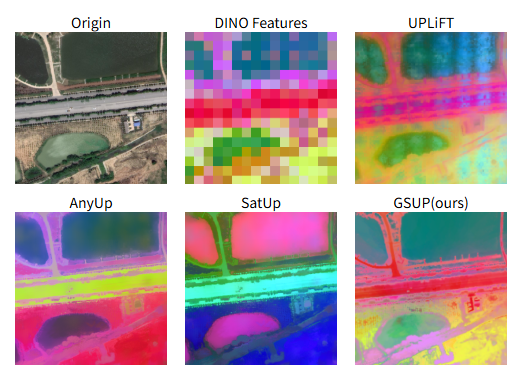}
    \caption{Visualization of feature reconstruction across different upsampling methods.}
    \label{fig:e}
\end{figure}

Table~2 show that GSUP achieves slightly lower accuracy than the pre‑trained optimal model, yet the gap is acceptable—reflecting a trade‑off between performance and efficiency. Its core strength lies in being dataset‑agnostic and pre‑training‑free: unlike AnyUp or SatUp, which learn parameters on large datasets, GSUP optimizes only Gaussian splatting parameters per test image (10 SGD iterations). This yields a good balance among deployment flexibility, theoretically maximal cross‑domain generalization, and cost, suitable for remote sensing with scarce annotations and high variability. Also note that FeatUp needs local CUDA compilation, AnyUp requires NATTEN, while GSUP needs no pre‑training, offering better portability with minimal modifications.

\subsection{Ablations}
To validate the effectiveness of each of the four proposed modules in adapting DINO.txt for training‑free remote sensing open‑vocabulary segmentation, we conduct ablation studies on the UDD5 dataset.
\begin{figure}[htbp]
    \centering
    \includegraphics[width=1\linewidth]{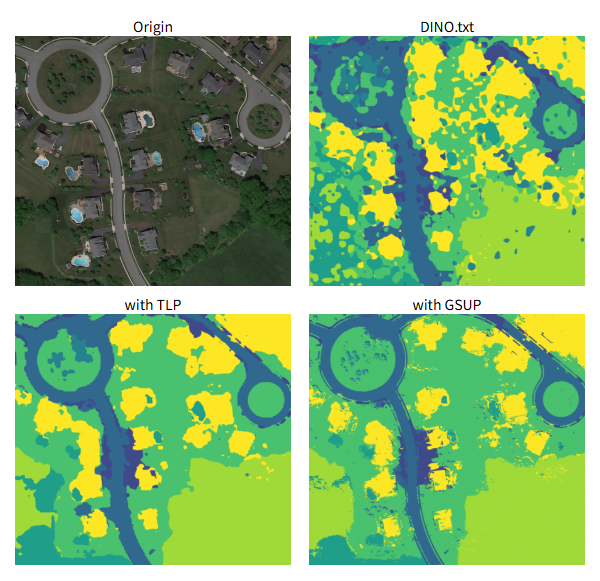}
    \caption{The features extracted by DINO.txt are first enhanced and denoised through Text‑aware Laplacian Propagation (TLP), and then reconstructed into pixel‑level high‑resolution representations via RGB‑guided Gaussian Splatting Upsampling (GSUP), ultimately yielding the final semantic segmentation predictions.}
    \label{fig:placeholder}
\end{figure}
\begin{table}[htbp]
\centering
\label{tab:ablation}
\begin{tabular}{l c c}
  \toprule
  \textbf{Method} & \textbf{Base mIoU} & \textbf{$\Delta$} \\
  \midrule
  w/o all (baseline) & 32.4 & 0 \\
  w/ Sliding Window & 36.1 & +3.7 \\
  w/ Synonym Aggregation & 37.7 & +1.6 \\
  w/ Laplacian Propagation (TLP) & 40.7 & +3.0 \\
  w/ Gaussian Splatting (GSUP) & 42.9 & +2.2 \\
  \bottomrule
\end{tabular}
\caption{Ablation study of the proposed modules on the UDD5 dataset.}

\end{table}

Table~3 shows that sliding window and synonym aggregation provide foundational improvements, increasing the mIoU from 32.4\% to 36.1\% and 37.7\%, respectively. These gains demonstrate their effectiveness in handling large-scale remote sensing images and alleviating the text-image matching ambiguity of DINO.txt. The proposed core modules, TLP and GSUP, further improve the performance to 40.7\% and 42.9\%, corresponding to absolute gains of +8.3\% and +10.5\% over the baseline, respectively. TLP performs text-guided diffusion to enhance semantic consistency, suppressing noisy predictions while preserving category boundaries. GSUP addresses the spatial resolution limitation of DINOv3 patch features by reconstructing high-resolution semantic representations through RGB-guided anisotropic Gaussian aggregation with test-time optimization. The complementary effects of semantic refinement and resolution recovery lead to the best performance when all modules are combined.
\subsection{Generalizability of different backbones}
\begin{table}[htbp]
\centering
\label{tab:ablation}
\begin{tabular}{l c c}
  \toprule
  \textbf{Backbone} & \textbf{Base mIoU} & \textbf{$\Delta$} \\
  \midrule
  \textbf{DINO.txt} & 37.7 & - \\
  + TLP & 40.7 & +3.0 \\
  + GSUP & 42.9 & +2.2 \\
  \textbf{ClearCLIP} & 38.2 & - \\
  + TLP & 40.8 & +2.6 \\
  + GSUP & 42.8 & +2.0 \\
  \bottomrule
\end{tabular}
\caption{Replace different backbones on the UDD5 dataset.}

\end{table}

As shown in Table~4, we further investigate the generalizability of TLP and GSUP. While SegEarth-OV incorporates a featUp upsampling module pretrained on Million-AID\citep{Long2021DiRS} dataset, which theoretically endows it with appreciable generalization capability, both TLP and GSUP require no dataset-specific pretraining at all. This allows them to achieve maximal generalization, effectively enabling \textbf{Segment-Anything} performance, particularly on less commonly used datasets.

\section{Conclusion}

DinoSplat‑OV is the first \textbf{training‑free} remote sensing open‑vocabulary segmentation framework built on the DINOv3 text encoder (DINO.txt). It requires no training or fine‑tuning, relying on two core inference‑time modules: Text‑aware Graph Laplacian Propagation (TLP) to align coarse features with semantic priors, and 2D Gaussian Splatting Upsampling (GSUP) to reconstruct low‑resolution features into high‑precision pixel‑level predictions. Extensive experiments on UDD5, DOTA, LoveDA, Vaihingen and other benchmarks show that DinoSplat‑OV achieves competitive or even superior performance over state‑of‑the‑art training‑free methods, filling the gap of DINO‑series models in this task. Moreover, by transferring the explicit scene representation of 3D Gaussian splatting to 2D feature recovery for dense segmentation, this work offers a lightweight, pre‑training‑free alternative for high‑resolution feature reconstruction from frozen backbones, with generality extendable to other architectures like CLIP or SigLIP.

\nocite{*}
%%BIBENTRY-END:c:79%%
\bibliography{aaai2027}

@misc{simeoni2025dinov3,
  title={{DINOv3}},
  author={Sim{\'e}oni, Oriane and Vo, Huy V. and Seitzer, Maximilian and Baldassarre, Federico and Oquab, Maxime and Jose, Cijo and Khalidov, Vasil and Szafraniec, Marc and Yi, Seungeun and Ramamonjisoa, Micha{\"e}l and Massa, Francisco and Haziza, Daniel and Wehrstedt, Luca and Wang, Jianyuan and Darcet, Timoth{\'e}e and Moutakanni, Th{\'e}o and Sentana, Leonel and Roberts, Claire and Vedaldi, Andrea and Tolan, Jamie and Brandt, John and Couprie, Camille and Mairal, Julien and J{\'e}gou, Herv{\'e} and Labatut, Patrick and Bojanowski, Piotr},
  year={2025},
  eprint={2508.10104},
  archivePrefix={arXiv},
  primaryClass={cs.CV},
  url={https://arxiv.org/abs/2508.10104},
}

@article{jose2024dinov2,
  title={DINOv2 Meets Text: A Unified Framework for Image- and Pixel-Level Vision-Language Alignment},
  author={Jose, Cijo and Moutakanni, Th{\'e}o and Kang, Dahyun and Baldassarre, Federico and Darcet, Timoth{\'e}e and Xu, Hu and Li, Daniel and Szafraniec, Marc and Ramamonjisoa, Micha{\"e}l and Oquab, Maxime and Sim{\'e}oni, Oriane and Vo, Huy V. and Labatut, Patrick and Bojanowski, Piotr},
  journal={arXiv preprint arXiv:2412.16334},
  year={2024},
  eprint={2412.16334},
  archivePrefix={arXiv},
  primaryClass={cs.CV},
  doi={10.48550/arXiv.2412.16334}
}

@article{wang2023sclip,
  title={SCLIP: Rethinking Self-Attention for Dense Vision-Language Inference},
  author={Wang, Feng and Mei, Jieru and Yuille, Alan},
  journal={arXiv preprint arXiv:2312.01597},
  year={2023}
}

@article{wysoczanska2023clipdino,
        title={CLIP-DINOiser: Teaching CLIP a few DINO tricks for open-vocabulary semantic segmentation},
        author={Wysocza{\'n}ska, Monika and Sim{\'e}oni, Oriane and Ramamonjisoa, Micha{\"e}l and Bursuc, Andrei and Trzci{\'n}ski, Tomasz and P{\'e}rez, Patrick},
        journal={ECCV},
        year={2024}
}

@inproceedings{lan2024clearclip,
  title={Clearclip: Decomposing clip representations for dense vision-language inference},
  author={Lan, Mengcheng and Chen, Chaofeng and Ke, Yiping and Wang, Xinjiang and Feng, Litong and Zhang, Wayne},
  booktitle={European Conference on Computer Vision},
  pages={143--160},
  year={2024},
  organization={Springer}
}

@article{cao2025open,
  title={Open-Vocabulary High-Resolution Remote Sensing Image Semantic Segmentation},
  author={Cao, Qinglong and Chen, Yuntian and Ma, Chao and Yang, Xiaokang},
  journal={IEEE Transactions on Geoscience and Remote Sensing},
  year={2025},
  publisher={IEEE}
}

@inproceedings{ye2025GSNet,
  title={Towards Open-Vocabulary Remote Sensing Image Semantic Segmentation},
  author={Ye, Chengyang and Zhuge, Yunzhi and Zhang, Pingping},
  booktitle={Proceedings of the AAAI Conference on Artificial Intelligence},
  year={2025}
}

@InProceedings{stojnic2025_lposs,
    author    = {Stojni\'c, Vladan and Kalantidis, Yannis and Matas, Ji\v{r}\'i  and Tolias, Giorgos},
    title     = {LPOSS: Label Propagation Over Patches and Pixels for Open-vocabulary Semantic Segmentation},
    booktitle = {Proceedings of the IEEE/CVF Conference on Computer Vision and Pattern Recognition (CVPR)},
    year      = {2025}
}

@inproceedings{lan2024proxyclip,
  title={Proxyclip: Proxy attention improves clip for open-vocabulary segmentation},
  author={Lan, Mengcheng and Chen, Chaofeng and Ke, Yiping and Wang, Xinjiang and Feng, Litong and Zhang, Wayne},
  booktitle={European Conference on Computer Vision},
  pages={70--88},
  year={2024},
  organization={Springer}
}

@inproceedings{li2025segearthov,
  title={Segearth-ov: Towards training-free open-vocabulary segmentation for remote sensing images},
  author={Li, Kaiyu and Liu, Ruixun and Cao, Xiangyong and Bai, Xueru and Zhou, Feng and Meng, Deyu and Wang, Zhi},
  booktitle={Proceedings of the Computer Vision and Pattern Recognition Conference},
  pages={10545--10556},
  year={2025}
}

@inproceedings{wimmer2026anyup,
    title={AnyUp: Universal Feature Upsampling},
    author={Wimmer, Thomas and Truong, Prune and Rakotosaona, Marie-Julie and Oechsle, Michael and Tombari, Federico and Schiele, Bernt and Lenssen, Jan Eric},
    booktitle={Proceedings of the International Conference on Learning Representations ({ICLR})},
    year={2026}
}

@misc{chambon2025nafzeroshotfeatureupsampling,
      title={NAF: Zero-Shot Feature Upsampling via Neighborhood Attention Filtering}, 
      author={Loick Chambon and Paul Couairon and Eloi Zablocki and Alexandre Boulch and Nicolas Thome and Matthieu Cord},
      year={2025},
      url={https://arxiv.org/abs/2511.18452}, 
}

@inproceedings{hajimiri2025naclip,
  title={Pay Attention to Your Neighbours: Training-Free Open-Vocabulary Semantic Segmentation},
  author={Hajimiri, Sina and Ben Ayed, Ismail and Dolz, Jose},
  year={2025},
  booktitle={Proceedings of the IEEE/CVF Winter Conference on Applications of Computer Vision},
}

@inproceedings{
    fu2024featup,
    title={FeatUp: A Model-Agnostic Framework for Features at Any Resolution},
    author={Stephanie Fu and Mark Hamilton and Laura E. Brandt and Axel Feldmann and Zhoutong Zhang and William T. Freeman},
    booktitle={The Twelfth International Conference on Learning Representations},
    year={2024},
    url={https://openreview.net/forum?id=GkJiNn2QDF}
}

@Article{kerbl3Dgaussians,
      author       = {Kerbl, Bernhard and Kopanas, Georgios and Leimk{\"u}hler, Thomas and Drettakis, George},
      title        = {3D Gaussian Splatting for Real-Time Radiance Field Rendering},
      journal      = {ACM Transactions on Graphics},
      number       = {4},
      volume       = {42},
      month        = {July},
      year         = {2023},
      url          = {https://repo-sam.inria.fr/fungraph/3d-gaussian-splatting/}
}

@inproceedings{lee2026glaclip,
  title={Looking Beyond the Window: Global-Local Aligned CLIP for Training-free Open-Vocabulary Semantic Segmentation},
  author={Lee, ByeongCheol and Seong, Hyun Seok and Hyun, Sangeek and Park, Gilhan and Moon, WonJun and Heo, Jae-Pil},
  booktitle={Proceedings of the IEEE/CVF Conference on Computer Vision and Pattern Recognition (CVPR)},
  year={2026}
}

@article{pei2026pearl,
  title={PEARL: Geometry Aligns Semantics for Training-Free Open-Vocabulary Semantic Segmentation},
  author={Pei, Gensheng and Jiang, Xiruo and Cai, Xinhao and Chen, Tao and Yao, Yazhou and Jeon, Byeungwoo},
  year={2026},
  journal={arXiv preprint arXiv:2603.21528},
}

@article{li2025segearthov3,
  title={SegEarth-OV3: Exploring SAM 3 for Open-Vocabulary Semantic Segmentation in Remote Sensing Images},
  author={Li, Kaiyu and Zhang, Shengqi and Wang, Yujie and Deng, Yupeng and Wang, Zhi and Meng, Deyu and Cao, Xiangyong},
  journal={arXiv preprint arXiv:2512.08730},
  year={2025}
}

@misc{cho2024catseg,
      title={CAT-Seg: Cost Aggregation for Open-Vocabulary Semantic Segmentation}, 
      author={Seokju Cho and Heeseong Shin and Sunghwan Hong and Anurag Arnab and Paul Hongsuck Seo and Seungryong Kim},
      year={2024},
      eprint={2303.11797},
      archivePrefix={arXiv},
      primaryClass={cs.CV}
}

@inproceedings{4e732ced,
         author = {Wang, Junjue and Zheng, Zhuo and Ma, Ailong and Lu, Xiaoyan and Zhong, Yanfei},
         booktitle = {Proceedings of the Neural Information Processing Systems Track on Datasets and Benchmarks},
         editor = {J. Vanschoren and S. Yeung},
         pages = {},
         publisher = {Curran Associates, Inc.},
         title = {LoveDA: A Remote Sensing Land-Cover Dataset for Domain Adaptive Semantic Segmentation},
         url = {https://datasets-benchmarks-proceedings.neurips.cc/paper_files/paper/2021/file/4e732ced3463d06de0ca9a15b6153677-Paper-round2.pdf},
         volume = {1},
         year = {2021}
}

@InProceedings{Xia_2018_CVPR,
author = {Xia, Gui-Song and Bai, Xiang and Ding, Jian and Zhu, Zhen and Belongie, Serge and Luo, Jiebo and Datcu, Mihai and Pelillo, Marcello and Zhang, Liangpei},
title = {DOTA: A Large-Scale Dataset for Object Detection in Aerial Images},
booktitle = {The IEEE Conference on Computer Vision and Pattern Recognition (CVPR)},
month = {June},
year = {2018}
}

@inproceedings{chen2018large,
  title={Large-scale structure from motion with semantic constraints of aerial images},
  author={Chen, Yu and Wang, Yao and Lu, Peng and Chen, Yisong and Wang, Guoping},
  booktitle={Chinese Conference on Pattern Recognition and Computer Vision (PRCV)},
  year={2018},
  organization={Springer}
}

@inproceedings{SAMRS,
 author = {Wang, Di and Zhang, Jing and Du, Bo and Xu, Minqiang and Liu, Lin and Tao, Dacheng and Zhang, Liangpei},
 booktitle = {Advances in Neural Information Processing Systems},
 pages = {8815--8827},
 title = {SAMRS: Scaling-up Remote Sensing Segmentation Dataset with Segment Anything Model},
 volume = {36},
 year = {2023}
}

@inproceedings{chen2025feat2gs,
  title={Feat2gs: Probing visual foundation models with gaussian splatting},
  author={Chen, Yue and Chen, Xingyu and Chen, Anpei and Pons-Moll, Gerard and Xiu, Yuliang},
  booktitle={Proceedings of the Computer Vision and Pattern Recognition Conference},
  pages={6348--6361},
  year={2025}
}

@article{walmer2026uplift,
  title={UPLiFT: Efficient Pixel-Dense Feature Upsampling with Local Attenders},
  author={Walmer, Matthew and Suri, Saksham and Aggarwal, Anirud and Shrivastava, Abhinav},
  journal={arXiv preprint arXiv:2601.17950},
  year={2026}
}

@article{rotinwa2012benchmark,
  title={ISPRS 2D Semantic Labeling Contest},
  author={Niemeyer, J. and Rottensteiner, F. and Soergel, U.},
  journal={ISPRS Annals of Photogrammetry, Remote Sensing and Spatial Information Sciences},
  volume={I-3},
  pages={293--298},
  year={2012}
}

@article{Long2021DiRS,
title={On Creating Benchmark Dataset for Aerial Image Interpretation: Reviews, Guidances and Million-AID},
author={Yang Long and Gui-Song Xia and Shengyang Li and Wen Yang and Michael Ying Yang and Xiao Xiang Zhu and Liangpei Zhang and Deren Li},
journal={IEEE Journal of Selected Topics in Applied Earth Observations and Remote Sensing},
year={2021},
volume={14},
pages={4205-4230}
}


\end{document}